# Vehicle in Virtual Environment (VVE) Method

Sukru Yaren Gelbal, Bilin Aksun-Guvenc, Levent Guvenc
Automated Driving Lab, Ohio State University

## Abstract

Autonomous vehicle (AV) algorithms need to be tested extensively in order to make sure the vehicle and the passengers will be safe while using it after the implementation. Testing these algorithms in real world create another important safety critical point. Real world testing is also subjected to limitations such as logistic limitations to carry or drive the vehicle to a certain location. For this purpose, hardware in the loop (HIL) simulations as well as virtual environments such as CARLA and LG SVL are used widely. This paper discusses a method that combines the real vehicle with the virtual world, called vehicle in virtual environment (VVE). This method projects the vehicle location and heading into a virtual world for desired testing, and transfers back the information from sensors in the virtual world to the vehicle. As a result, while vehicle is moving in the real world, it simultaneously moves in the virtual world and obtains the situational awareness via multiple virtual sensors. This would allow testing in a safe environment with the real vehicle while providing some additional benefits on vehicle dynamics fidelity, logistics limitations and passenger experience testing. The paper also demonstrates an example case study where path following and the virtual sensors are utilized to test a radar based stopping algorithm.

## Introduction

The literature is full of papers on active safety and Advanced Driver Assistance Systems (ADAS) like references [1], [2], [3], [4], [5]; other references like [6], [7] on connected vehicles (CV) and autonomous vehicles (AV) along with references on robust and fuel optimal controls and automated driving algorithms like [8], [9], [10], [11], [12]; those on AV shuttles and taxis like [13], [14]; those associated with fuel/energy economy and platooning like [15], [16], [17], [18], [19], [20], [21], [22], [23], [24], [25] and path planning and tracking control of AVs like [26]. All of these ADAS, CV and AV algorithms need to be realistically tested in a safe and time efficient manner as opposed to the current approach of testing on public roads.

The Testing stage for automated vehicle algorithms and controllers is a critical stage where many factors can be in play for the results to be inaccurate or misleading. Moreover, the availability of the test scenarios, cost, logistics and most importantly, the safety factor, depends on the mode of testing. Wide variety of testing modes are utilized for automated vehicles in recent years. One of these testing modes that is also widely used is simulation based testing or software in the loop (SIL) [27]. The simulation environment provides a large amount of freedom, scalability, efficiency and gets rid of safety concerns involving some dangerous test scenarios. This mode of testing is getting more popular especially with the respective tools becoming more capable and accessible such as CARLA [28] and LG SVL [29]. Another widely used mode is the HIL simulations [30]. These incorporate real hardware such as real sensors, communication modules, control or processing units into the simulation environment, in order to incorporate the factors which come along with real hardware. Finally, vehicle in the loop (VIL) [31] is a mode of testing with the whole vehicle is incorporated into the testing, with benefits of cost effectiveness, logistics, real hardware, real control, actuation and vehicle dynamics.

VVE is a concept similar to VIL, combining the simulation environment with hardware where hardware incorporated is the whole vehicle, hardware processors, controllers and actuators. In addition to using the vehicle inside the loop, all of the information related to situational awareness is obtained from the simulation environment. VVE allows the real vehicle to be immersed into a virtual world where wide range of scenarios and sensors are easily accessible, much safer and cost effective to test. With this immersion, vehicle is able to utilize these advantages of the virtual world while bringing along its automated driving algorithms and controllers carried as hardware and software, as well as the actuation and vehicle dynamics. Moreover, a passenger can also be carried in the real vehicle, can even be connected to a virtual reality headset, to test the passenger experience and comfort for the automated driving. The combination of these advantages from each world makes the VVE a very interesting mode of evaluation and algorithm development for autonomous vehicles.

The rest of the paper is organized as follows. In the following section, VVE structure is explained in detail with components and communication between them. Followed by discussion about the simulation environment. Afterwards, paper demonstrates this concept on a real world implementation with automated path following. Finally, paper concludes with conclusion and future work section.

## VVE Structure

In order to immerse the real vehicle into virtual world, the concept of VVE requires the information of real world to be translated into the virtual world, and vice-versa. This translation is both aimed to match the real world and virtual world with respect to ego vehicle motion as accurately as possible, and to utilize the virtual sensors in the simulation to test automated driving algorithms carried out by the real autonomous vehicle. Overview of the created VVE structure in Figure 1 follows this principle.

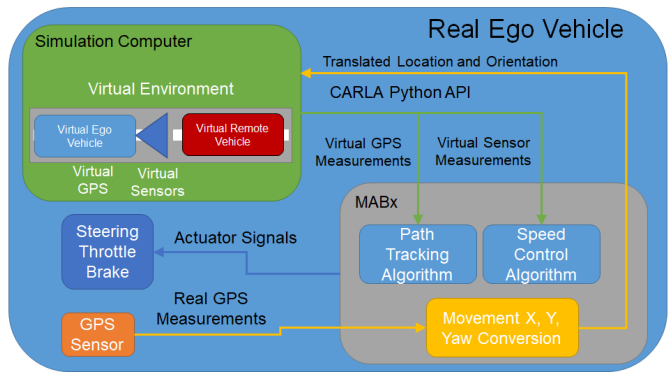

Figure 1. Diagram of VVE structure inside the vehicle.

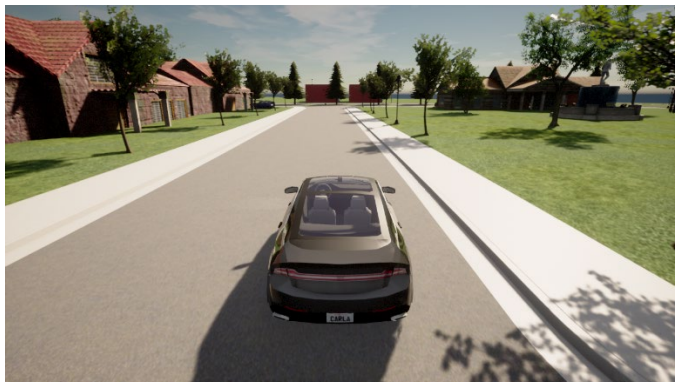

Figure 2. Autonomous vehicle driving in virtual Linden.

In the virtual world, virtual ego vehicle has to be moved in such a way that it would match the real world vehicle movement, which would require location and orientation information from the real world vehicle. On the other side, real world vehicle requires sensor information from the virtual ego vehicle in order to be immersed into the virtual world. This communication was coded by authors for VVE utilizing the CARLA Python API along with other simulation functionality, as shown in Figure 1.

For our use case, where automated driving will be utilized in the real world as well, we transfer this information through a dSpace MicroAutoBox (MABx). MABx acts as a lower level controller, actuating the real world vehicle steering, throttle and brake through controller area network (CAN) bus, while at the same time processing the sensor information and making decisions on how or when to drive or stop the vehicle. As seen in the diagram, containing the path following and speed control algorithms, MABx plays the role of the brain of the vehicle that is connected and immersed into the virtual world. As a result, vehicle uses the virtual global positioning sensor (GPS) measurements for automated path following, while using other virtual sensors (can be camera, lidar, radar) to detect other road users and the obstacles, in order to make correct decisions accordingly.

The translation of the real-world motion into the simulation environment is also done by MABx in our structure. Translation is done with a GPS coordinates (degree) to meters conversion based on the virtually matched starting location.

## Simulation Environment in CARLA

Automated Driving Lab has been working on creating realistic virtual environments for testing automated driving functions. The most recent work [32] was related to EasyMile deployment in Linden, Ohio [33] for Linden LEAP project. This virtual environment was also created in Unreal Engine, utilizing RoadRunner for the road structure along with available 3D assets to replicate houses and trees. With the graphical fidelity of Unreal Engine, CARLA also allows us to represent both the environment and road users as realistically as possible from sensor perspective. As a result, these virtual environments replicated from real world locations, allow us to conduct automated driving tests in this specific area much safer and more efficiently with wide array of available sensors from CARLA simulator. A screenshot can be seen in Figure 2 where autonomous vehicle is driving on a street at our virtual Linden.

In our VVE setup, the simulation environment runs in a separate simulation computer discussed in the previous section. This allows the computer to focus on simulating and rendering the environment and data acquisition from the virtual sensors, which require significant amount of processing power overall. CARLA, with flexibility obtained by utilizing Unreal Engine, can incorporate wide variety of road users, environments and driving scenarios including independent agents for realistic traffic.

## Real World Testing and Demonstration

For real-world testing and demonstration of VVE, our Ford Fusion autonomous vehicle was utilized with its hardware and software. The vehicle is drive-by-wire and has automated driving capabilities [34]. A picture of the vehicle and the hardware components in the trunk for implementation of VVE can be seen in Figure 3 below.

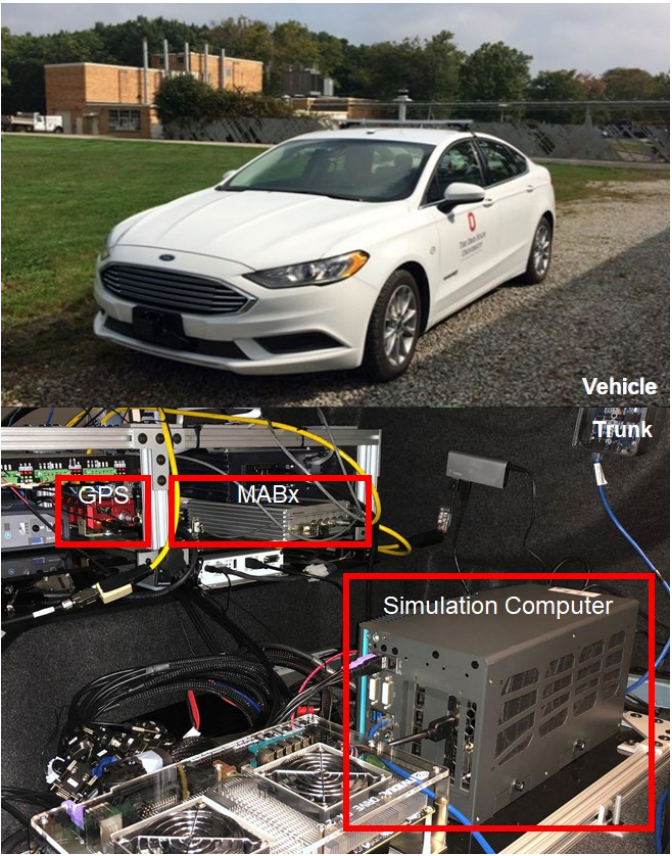

Figure 3. Ford Fusion autonomous vehicle and VVE hardware components in the trunk.

As discussed in previous sections, CARLA based simulation environment was implemented into the simulation computer shown in Figure 3. The communication was coded between MABx and this computer, along with required automated driving algorithms, decision making and control carried out by MABx itself. An accurate GPS sensor is beneficial for precise measurement of the real world vehicle location, in order to translate into the virtual world. Our vehicle carries an OXTS brand GPS with real-time kinematics (RTK) corrections for up to a few centimeters of accuracy.

After the hardware and software implementation of the VVE structure, an automated driving scenario was created involving virtual sensors. In this driving scenario, while the ego vehicle is doing automated path following, it comes across a parked vehicle in the middle of the road and has to stop and wait. The vehicle also carries a virtual radar sensor for detecting other vehicles and obstacles. The scenario can be roughly illustrated as in Figure 4.

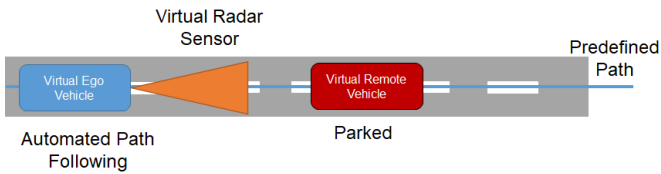

Figure 4. Illustration of the test scenario.

In this scenario, both virtual GPS measurements and the virtual radar measurements has to be conveyed to the real vehicle, which will be over the ethernet connection in our case. After real vehicle picks up the information via MABx, it will be processed according to path following and decision making algorithm. Then, actuation commands will be obtained and applied to the real vehicle through CAN bus to conduct automated path following and correctly stop at a safe distance to the parked vehicle.

In order to create the predefined path for the virtual vehicle, GPS data collection was done in the virtual world by driving the vehicle manually on a street at Linden. The collected GPS points can be seen in Figure 5.

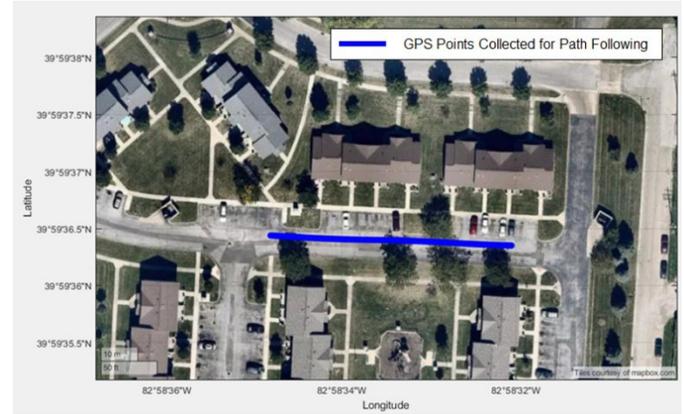

Figure 5. GPS data collected from simulation environment plotted on a satellite image from Linden.

The points were used to generate a path for the vehicle to follow autonomously. Finally, the test was conducted in a parking lot with free space enough to complete the testing and demonstration of this scenario. The pictures from the test can be seen in Figure 6 where the video recording can be accessed through [35]. Pictures and the video feature a view from within the vehicle where both the real world testing environment and 3D rendering of the simulation environment is visible. Real vehicle motion can be observed from the windshield where virtual environment can be observed from the small screen on the dashboard.

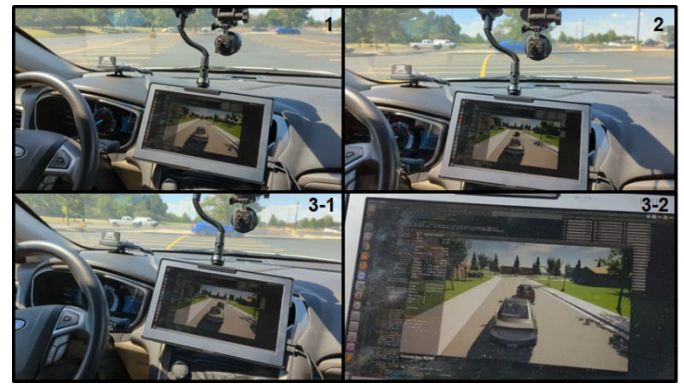

Figure 6. Pictures from 3 phases of the testing, start, path following and stop.

Figure 6 demonstrates a few pictures from the video recording during the testing. These pictures include a few phases of the test, where picture 1 is the start of the test, picture 2 is during the path following, picture 3-1 is the part where vehicle stops after detecting the virtual parked vehicle and picture 3-2 is zoomed in view of the simulation screen to show the final state of the virtual vehicle. When the pictures and video are observed for this demonstration scenario, it can be seen that the real vehicle is successfully doing path-following with a pre-defined path from a virtual environment while obtaining and processing virtual sensor information for decision making. Utilizing the virtual radar sensor, vehicle is also successfully able to slow down and stop at a safe distance at the end, autonomously. This stop behavior is triggered by virtual radar sensor measurements due to the virtual parked vehicle in the scenario. Through the testing, real vehicle can be observed from the pictures and the video as moving in the free space of the testing area.

## Conclusions and Future Work

VVE is a mode of testing for automated vehicles that immerses the vehicle into a realistic virtual world. Situational awareness is achieved by virtual sensors in the simulation environment, while real vehicle is actuated and controlled by its own hardware and automated driving algorithms. This mode combines the advantages of the virtual world such as cost efficiency, logistical time efficiency, safety, test scenario flexibility with the advantages of the real world such as, incorporation of real vehicle, hardware and software components from the automated driving architecture.

VVE was discussed in the paper in terms of the concept, hardware and software implementation structure. Implementation for a real vehicle was also discussed along with a demonstration scenario where vehicle concludes the testing successfully. For the future work, the VVE structure is capable of incorporating more real world components. One component to include would be a passenger, who can participate in the testing for comfort. With augmented reality googles or a virtual reality headset, pedestrian can see the virtual world while experiencing the physical effects during autonomous driving. Another component to include would be a real vulnerable road user (VRU). The VRU can move at a distant safe location while the information is transferred to virtual world via communication. This would allow collision avoidance testing with real human behavior.

## References

1. Aksun-Guvenc, B., Guvenc, L., Ozturk, E.S., Yigit, T., "Model Regulator Based Individual Wheel Braking Control," IEEE Conference on Control Applications, Istanbul, June 23-25, 2003.
2. Aksun-Guvenc, B., Guvenc, L., "The Limited Integrator Model Regulator and its Use in Vehicle Steering Control," *Turkish Journal of Engineering and Environmental Sciences*, pp. 473-482, 2002.
3. Oncu, S., Karaman, S., Guvenc, L., Ersolmaz, S.S., Ozturk, E.S., Cetin, A.E., Sinal, M., "Steer-by-Wire Control of a Light Commercial Vehicle Using a Hardware-in-the-Loop Setup," IEEE Intelligent Vehicles Symposium, June 13-15, pp. 852-859, 2007.
4. Emirler, M.T., Kahraman, K., Senturk, M., Aksun-Guvenc, B., Guvenc, L., Efendioglu, B., "Two Different Approaches for Lateral Stability of Fully Electric Vehicles," *International Journal of Automotive Technology*, Vol. 16, Issue 2, pp. 317-328, 2015.
5. Kural, E., Aksun-Guvenc, B., "Model Predictive Adaptive Cruise Control," IEEE International Conference on Systems, Man, and Cybernetics, Istanbul, Turkey, October 10-13, 2010.
6. Guvenc, L., Aksun-Guvenc, B., Demirel, B., Emirler, M.T., *Control of Mechatronic Systems*, the IET, London, ISBN: 978-1-78561-144-5, 2017.
7. Boyali A., Guvenc, L., "Real-Time Controller Design for a Parallel Hybrid Electric Vehicle Using Neuro-Dynamic Programming Method," IEEE Systems, Man and Cybernetics, Istanbul, October 10-13, pp. 4318-4324, 2010.
8. Zhou, H., Jia, F., Jing, H., Liu, Z., Guvenc, L., 2018, "Coordinated Longitudinal and Lateral Motion Control for Four Wheel Independent Motor-Drive Electric Vehicle," *IEEE Transactions on Vehicular Technology*, Vol. 67, No 5, pp. 3782-379.
9. Wang, H., Tota, A., Aksun-Guvenc, B., Guvenc, L., 2018, "Real Time Implementation of Socially Acceptable Collision Avoidance of a Low Speed Autonomous Shuttle Using the Elastic Band Method," *IFAC Mechatronics Journal*, Volume 50, April 2018, pp. 341-355.
10. Bowen, W., Gelbal, S.Y., Aksun-Güvenç, B., Guvenc, L., 2018, "Localization and Perception for Control and Decision Making of a Low Speed Autonomous Shuttle in a Campus Pilot Deployment," *SAE International Journal of Connected and Automated Vehicles*, doi: 10.4271/12-01-02-0003, Vol. 1, Issue 2, pp. 53-66.
11. Emirler, M.T., Uygan, I.M.C., Aksun-Guvenc, B., Guvenc, L., 2014, "Robust PID Steering Control in Parameter Space for Highly Automated Driving," *International Journal of Vehicular Technology*, Vol. 2014, Article ID 259465.
12. Emirler, M.T., Wang, H., Aksun-Guvenc, B., Guvenc, L., 2015, "Automated Robust Path Following Control based on Calculation of Lateral Deviation and Yaw Angle Error," ASME Dynamic Systems and Control Conference, DSC 2015, October 28-30, Columbus, Ohio, U.S.
13. Gelbal, S.Y., Aksun-Guvenc, B., Guvenc, L., 2019, SmartShuttle: Model Based Design and Evaluation of Automated On-Demand Shuttles for Solving the First-Mile and Last-Mile Problem in a Smart City, Final Research Report, arXiv:2012.12431v1 [cs.RO].
14. Guvenc, L., Aksun-Guvenc, B., Li, X., Arul Doss, A.C., Meneses-Cime, K.M., Gelbal, S.Y., 2019, Simulation Environment for Safety Assessment of CEAV Deployment in Linden, Final Research Report, Smart Columbus Demonstration Program – Smart City Challenge Project, arXiv:2012.10498 [cs.RO].
15. Kavas-Torris, O., Lackey, N.A., Guvenc, L., 2021, "Simulating Autonomous Vehicles in a Microscopic Traffic Simulator to Investigate the Effects of Autonomous Vehicles on Roadway Mobility," *International Journal of Automotive Technology*, Vol. 22, No. 4, pp. 713–733.
16. Kavas-Torris, O., Cantas, M.R., Gelbal, S.Y., Aksun-Guvenc, B., Guvenc, L., 2020, "Fuel Economy Benefit Analysis of Pass-at-Green (PaG) V2I Application on Urban Routes with STOP Signs," Special Issue on Safety and Standards for CAV, *International Journal of Vehicle Design*, Vol. 83, No. 2/3/4, pp. 258-279.
17. Yang, Y., Ma, F., Wang, J., Zhu, S., Gelbal, S.Y., Kavas-Torris, O., Aksun-Guvenc, B., Guvenc, L., 2020, "Cooperative Ecological Cruising Using Hierarchical Control Strategy with Optimal Sustainable Performance for Connected Automated Vehicles on Varying Road Conditions," *Journal of Cleaner Production*, Vol. 275, doi.org/10.1016/j.jclepro.2020.123056.
18. Kavas-Torris, O., Guvenc, L., "A Comprehensive Eco-Driving Strategy for Connected and Autonomous Vehicles (CAVs) with Microscopic Traffic Simulation Testing Evaluation," arXiv:2206.08306v1 [eess.SY].
19. Kavas-Torris, O., Cantas, M.R., Meneses-Cime, K., Aksun-Guvenc, B., Guvenc, L., 2020, "The Effects of Varying Penetration Rates of L4-L5 Autonomous Vehicles on Fuel Efficiency and Mobility of Traffic Networks," WCX20: SAE World Congress Experience, April 21-23, Detroit, Michigan, Session IDM300 Modelling, Simulation and Testing, SAE Paper Number: 2020-01-0137.
20. Ma, F., Wang, J., Zhu, S., Gelbal, S.Y., Yu, Y., Aksun-Guvenc, B., Guvenc, L., 2020, "Distributed Control of Cooperative

Vehicular Platoon with Nonideal Communication Condition," *IEEE Transactions on Vehicular Technology*, Vol. 69, Issue 8, pp. 8207-8220, doi:10.1109/TVT.2020.2997767.
21. Cebi, A., Guvenc, L., Demirci, M., Kaplan Karadeniz, C., Kanar, K., Guraslan, E., 2005, "A Low Cost, Portable Engine ECU Hardware-In-The-Loop Test System," Mini Track of Automotive Control, *IEEE International Symposium on Industrial Electronics Conference*, Dubrovnik, June 20-23.
22. S Zhu, B Aksun-Guvenc, 2020, "Trajectory planning of autonomous vehicles based on parameterized control optimization in dynamic on-road environments," *Journal of Intelligent & Robotic Systems*, 100 (3), 1055-1067.
23. Emirler, M.T., Guvenc, L., Aksun-Guvenc, B., 2018, "Design and Evaluation of Robust Cooperative Adaptive Cruise Control Systems in Parameter Space," *International Journal of Automotive Technology*, Vol. 19, Issue 2, pp. 359-367.
24. Zhu, S., Gelbal, S.Y., Aksun-Guvenc, B., Guvenc, L., 2019, "Parameter-space Based Robust Gain-scheduling Design of Automated Vehicle Lateral Control," *IEEE Transactions on Vehicular Technology*, doi: 10.1109/TVT.2019.2937562, Vol. 68, Issue 10, pp. 9660-9671.
25. Gelbal, S.Y., Aksun-Guvenc, B., Guvenc, L., 2020, "Elastic Band Collision Avoidance of Low Speed Autonomous Shuttles with Pedestrians," *International Journal of Automotive Technology*, Vol. 21, No. 4, pp. 903-917.
26. Guvenc, L., Aksun-Guvenc, B., Zhu, S., Gelbal, S.Y., 2021, *Autonomous Road Vehicle Path Planning and Tracking Control*, Wiley / IEEE Press, Book Series on Control Systems Theory and Application, New York, ISBN: 978-1-119-74794-9.
27. M. F. S. Ahamed, G. Tewolde and J. Kwon, "Software-in-the-Loop Modeling and Simulation Framework for Autonomous Vehicles," in *IEEE International Conference on Electro/Information Technology (EIT)*, 2018.
28. "CARLA Simulator," [Online]. Available: https://carla.org/.
29. "LG SVL Simulator," [Online]. Available: https://www.svlsimulator.com/.
30. M. R. Cantas, O. Kavas, S. Tamilarasan, S. Y. Gelbal and L. Guvenc, "Use of Hardware in the Loop (HIL) Simulation for Developing Connected Autonomous Vehicle (CAV) Applications," in *SAE World Congress Experience*, 2019.
31. T. Bokc, M. Maurer and G. Farbe, "Validation of the Vehicle in the Loop (VIL); A milestone for the simulation of driver assistance systems," in *IEEE Intelligent Vehicles Symposium*, 2007.
32. L. Guvenc, B. Aksun-Guvenc, X. Li, A. C. A. Doss, K. Meneses-Cime and S. Y. Gelbal, "Simulation Environment for Safety Assessment of CEAV Deployment in Linden," in *https://arxiv.org/abs/2012.10498*, 2020.
33. EasyMile, "EZ10 Autonomous Shuttles deployed in Columbus Neighbourhood," 2020. [Online]. Available: https://easymile.com/news/ez10-autonomous-shuttles-deployed-columbus-neighbourhood.
34. S. Y. Gelbal, N. Chandramouli, H. Wang, B. Aksun-Guvenc and L. Guvenc, "A unified architecture for scalable and replicable autonomous shuttles in a smart city," in *IEEE International Conference on Systems, Man, and Cybernetics*, Banff, Canada, 2017.
35. ADL, "VVE Demonstration Video," 2022. [Online]. Available: https://buckeyemailosu-my.sharepoint.com/:v:/g/personal/gelbal_1_buckeyemail_osu_edu/ESc_hJtNjoVIpWUZ3VlI7HkB3GoZ3G1nhf9bNH_OR2NwuQ?e=CqT4i1.

## Contact Information

Sukru Yaren Gelbal
gelbal.1@osu.edu
Automated Driving Lab, Ohio State University
1320 Kinnear Rd., Columbus, OH, 43212

## Definitions/Abbreviations

| | |
|---|---|
| **AV** | Autonomous Vehicle |
| **HIL** | Hardware in the Loop |
| **VVE** | Vehicle in Virtual Environment |
| **SIL** | Software in the Loop |
| **VIL** | Vehicle in the Loop |
| **MABx** | dSpace MicroAutoBox |
| **GPS** | Global Positioning System |
| **CAN** | Controller Area Network |
| **RTK** | Real-Time Kinematics |
| **VRU** | Vulnerable Road User |